\documentclass{llncs}
\usepackage{epsf}

\usepackage{listings}
\usepackage{xspace}
\newcommand{\webquestions}{\textsc{WebQuestions}\xspace}
\newcommand{\geoquery}{\textsc{GeoQuery}\xspace}
\newcommand{\free}{\textsc{Free917}\xspace}
\newcommand{\sempre}{\textsc{Sempre}\xspace}
\newcommand{\parasempre}{\textsc{ParaSempre}\xspace}
\newcommand{\deepqa}{\textsc{DeepQA}\xspace}

\usepackage{csvsimple}

\begin{document}

\title{Simple, Fast Semantic Parsing with a Tensor Kernel}

\author{Daoud Clarke}
\institute{
  Department of Informatics\\
  University of Sussex\\
  Falmer, Brighton, UK.\\
  daoud.clarke@gmail.com
}

\maketitle

\begin{abstract}

We describe a simple approach to semantic parsing based on a tensor
product kernel. We extract two feature vectors: one for the query and
one for each candidate logical form. We then train a clasifier using
the tensor product of the two vectors. Using very simple features for
both, our system achieves an average F1 score of 40.1\% on the
\webquestions dataset. This is comparable to more complex systems but
is simpler to implement and runs faster.

\end{abstract}

\section{Introduction}

In recent years, the task of semantic parsing for querying large
databases has been studied. This task differs from early work in
semantic parsing in several ways:
\begin{itemize}
\item The databases being queried are typically several orders of
  magnitude larger, contain much more diverse content, and are less
  structured.
\item In standard semantic parsing approaches, the aim is to learn a
  logical form to represent a query. In recent approaches the goal is
  to find the correct answer (entity or set of entities in the
  database), with learning a logical form a potential byproduct.
\item Because of this, the datasets, which would have consisted of
  queries together with their corresponding logical forms, now may
  consist of the queries together with the desired correct answer.
\item The datasets themselves are much larger, and cover a more
  diverse range of entities, however there may be a lot of overlap in
  the \emph{type} of queries in the dataset.
\end{itemize}
We believe it is the last of these points that means that simple
techniques such as the one we present can work surprisingly well. For
example, the \webquestions dataset contains 83 questions containing
the term ``currency''; of these 79 are asking what the currency of a
particular country is. These 79 questions can be answered using the
same logical form template, thus a system only has to see the term
``currency'', and identify the correct country in the question to have
a very good chance of getting the answer correct.

Knowing this on its own is not enough to build an effective system
however. We still need to be able to somehow identify that it is this
particular term in the query that is associated with this logical
form. In this paper we demonstrate one way that this can be
achieved. We build on the paraphrasing approach of \cite{Berant:14} in
that we use a fixed set of templates to generate a set of candidate
logical forms to answer a given query and map each logical form to a
natural language expression, its \emph{canonical utterance}. Instead
of using a complex paraphrasing model however, we use tensor kernels
to find relationships between terms occuring in the query and in the
canonical utterance. The virtue of our approach is in its simplicity,
which both aids implementation and speeds up execution.

\begin{figure}

\noindent
\begin{minipage}[t]{.5\textwidth}
\begin{lstlisting}
what caused the asian currency crisis?
what countries use the euro as official currency?
what currency can you use in aruba?
what currency do i bring to cuba?
what currency do i need in cuba?
what currency do i need in egypt?
what currency do i take to turkey?
what currency do italy have?
what currency do mexico use?
what currency do the ukraine use?
what currency do they accept in kenya?
what currency do they use in qatar?
what currency do you use in costa rica?
what currency does brazil use?
what currency does greece use 2012?
what currency does greece use?
what currency does hungary have?
what currency does jamaica accept?
what currency does ontario canada use?
what currency does senegal use?
what currency does south africa have?
what currency does thailand accept?
what currency does thailand use?
what currency does the dominican republic?
what currency does turkey accept?
what currency in dominican republic should i bring?
what currency is best to take to dominican republic?
what currency is used in england 2012?
what currency is used in france before euro?
what currency is used in germany 2012?
what currency is used in hungary?
what currency is used in switzerland 2012?
what currency should i bring to italy?
what currency should i take to dubai?
what currency should i take to jamaica?
what currency should i take to mauritius?
what currency should you take to thailand?
what currency to take to side turkey?
what do you call russian currency?
what is australian currency?
what is currency in dominican republic?
what is currency in panama?
what is the best currency to take to egypt 2013?
\end{lstlisting}
\end{minipage}%
\begin{minipage}[t]{.5\textwidth}
\begin{lstlisting}
what is the currency in australia 2011?
what is the currency in croatia 2012?
what is the currency in england 2012?
what is the currency in france?
what is the currency in germany in 2010?
what is the currency in slovakia 2012?
what is the currency in the dominican republic 2010?
what is the currency in the dominican republic called?
what is the currency in the republic of congo?
what is the currency name of brazil?
what is the currency name of china?
what is the currency of germany in 2010?
what is the currency of mexico called?
what is the currency of spain called?
what is the currency of sweden called?
what is the currency used in brazil?
what is the currency used in tunisia?
what is the local currency in the dominican republic?
what is the money currency in guatemala?
what is the money currency in italy?
what is the money currency in switzerland?
what is the name of currency used in spain?
what is the official currency in france?
what kind of currency do they use in thailand?
what kind of currency does cuba use?
what kind of currency does greece have?
what kind of currency does jamaica use?
what kind of currency to bring to mexico?
what money currency does canada use?
what the currency in argentina?
what type of currency does brazil use?
what type of currency does egypt have?
what type of currency does the us have?
what type of currency is used in puerto rico?
what type of currency is used in the united kingdom?
what type of currency should i take to mexico?
what's sweden's currency?
what's the egyptian currency?
which country has adopted the euro as its currency ( 1 point )?
which country uses euro as its main currency?
\end{lstlisting}
\end{minipage}
\caption{Questions from the \webquestions dataset containing the term
  ``currency''.}
\label{fig:currency}
\end{figure}

\section{Background}

The task of semantic parsing initially focussed on fairly small
problems, such as the \geoquery dataset, which initially consisted of
250 queries \cite{Zelle:96} and was later extended to around 1000
queries \cite{Tang:01}. Approaches to this task included inductive
logic programming \cite{Zelle:96,Tang:01}, probabilistic grammar
induction \cite{Zettlemoyer:05,Kwiatkowksi:10}, synchronous grammars
\cite{Wong:07} and induction of latent logical forms \cite{Liang:11},
the current state of the art on this type of dataset.

More recently, attention has focussed on answering queries in much
larger domains, such as Freebase \cite{Bollacker:08}, which contains
at the time of writing of around 2.7 billion facts. There are two
datasets of queries for this database: \free consisting of 917
questions annotated with logical forms \cite{Cai:13}, and
\webquestions which consists of 5,810 question-answer pairs, with no
logical forms \cite{Berant:13}. Approaches to this task include schema
matching \cite{Cai:13}, inducing latent logical forms
\cite{Berant:13}, application of paraphrasing techniques
\cite{Berant:14,Wang:14}, information extraction \cite{Yao:14},
learning low dimensional embeddings of words and knowledge base
constituents \cite{Bordes:14} and application of logical reasoning in
conjunction with statistical techniques \cite{Wang:14}. Note that
most of these approaches do not require annotated logical forms, and
either induce logical forms when training using the given answers, or
bypass them altogether.

\subsection{Semantic Parsing via Paraphrasing}

The \parasempre system of \cite{Berant:14} is based on the idea of
generating a set of candidate logical forms from the query using a set
of templates. For example, the query \emph{Who did Brad Pitt play in
  Troy?} would generate the logical form

\begin{center}
\texttt{Character.(Actor.BraddPitt $\sqcap$ Film.Troy)}
\end{center}

\noindent
as well as many incorrect logical forms. These are built by finding
substrings of the query that approximately match Freebase entities and
then applying relations that match the type of the entity. Given a
logical form, a canonical utterance is generated, again using a set of
rules, which depend on the syntactic type of the description of the
entities.

To identify the most likely logical form given a query, a set of
features are extracted from the query, logical form and canonical
utterance:
\begin{itemize}
\item Features extracted from the logical form itself, such as the
  size of the denotation of a logical form, i.e.~the number of results
  returned when evaluating the logical form as a query on the
  database. This is important, since many incorrect logical forms have
  denotation zero; this feature acts as a filter removing these.
\item Features derived from an association model. This involves
  examining spans in the query and canonical utterance and looking for
  paraphrases between these spans. These paraphrases are derived from
  a large paraphrase corpus and WordNet \cite{Fellbaum:98}.
\item Features derived from a vector space model built using Word2Vec
  \cite{Mikolov:13}.
\end{itemize}

In an analysis on the development set of \webquestions, the authors
showed that removing the vector space model lead to a small drop in
performance, removing the asssociation model gave a larger drop, and
removing both of these halved the performance score.

\section{Tensor Kernerls for Semantic Parsing}

We know that simple patterns or occurrences in the query can be used
to identify a correct logical form with high probability, as with the
``currency'' example. We still need some way of identifying these
patterns and linking them up to appropriate logical forms. In this
section we discuss one approach for doing this.

Our goal is to learn a mapping from queries to logical forms. One way
of doing this to consider a fixed number of logical forms for each
query sentence, and train a classifier to choose the best logical form
given a sentence \cite{Berant:14}. In order to ues this approach, we
need a single feature vector for each pair of queries and logical
forms. Our proposal is to extract features for each query and logical
form indepdendently, and to take their tensor product as the combined
vector. Explicitly, let $Q$ be the set of all possible queries and
$\Lambda$ be the set of all possible logical forms. For each query
$q\in Q$ and logical form $\lambda\in \Lambda$, we represent the pair
$(q,\lambda)$ by the vector:
$$\phi(q, \lambda) = \phi_Q(q)\otimes\phi_\Lambda(\lambda)$$
where $\phi_Q$ and $\phi_\Lambda$ map queries and logical forms to a
vector space, i.e. perform feature extraction.

Whilst this could potentially be a large space, note that we can use
the kernel trick to avoid computing very large vectors, using a simple
identity of dot products on tensor spaces:
$$\phi(q_1, \lambda_1)\cdot\phi(q_2,\lambda_2) = (\phi_Q(q_1)\cdot\phi_Q(q_2))(\phi_\Lambda(\lambda_1)\cdot\phi_\Lambda(\lambda_2))$$
The advantage of using the tensor product is that it preserves all the
information of the original vectors, allowing us to learn how features
relating to queries map to features relating to logical forms.

More generally, instead of representing the query and logical form as
vectors directly, this can be done implicitly using kernels. For
example, we may use a string kernel $\kappa_1$ on $Q$ and a tree
kernel $\kappa_2$ on $\Lambda$, then define the kernel $\kappa(q,
\lambda) = \kappa_1(q)\kappa_2(\lambda)$ on $Q\times \Lambda$. This
idea is closely related to the Schur product kernel
\cite{Shawe-Taylor:04}.

It is worth noting at this point that, while what we really want is a
one-to-one mapping from queries to logical forms, the classifier
actually gives us a set of logical forms for each query: we simply ask
it to classify each pair $(q,\lambda)$. In a probabilistic approach,
such as logistic regression, we can choose the $\lambda$ for which the
classifier gives the highest probability for $(q, \lambda)$.

\subsection{Application to Semantic Parsing via Paraphrasing}

There are clearly many ways we could map queries and logical forms to
vectors. In this paper we will consider one simple approach in which
we use unigrams as the features for both the query and the canonical
utterance associated with the logical form. In this case, the tensor
product of the vectors corresponds directly to the cartesian product
of the unigrams derived from the query with those from the canonical
utterance.

Recall that given two vector spaces $U$ and $V$ of dimensionality $n$
and $m$, the tensor product space $U\otimes V$ has dimensionality
$nm$. If we have bases for $U$ and $V$, then we can construct a basis
for $U\otimes V$. For each pair of basis vectors $u$ and $v$ in $U$
and $V$ respectively, we take a single basis vector $u\otimes v \in
U\otimes V$. In our case, the dimensions of $U$ and $V$ correspond to
terms that can occur as unigram features in the query or canonical
utterance respectively. Thus each basis vector of $U\otimes V$
corresponds to a pair of unigram features.

As an example from the \webquestions dataset, consider the query,
\emph{What 5 countries border ethiopia?}, and the canonical utterance
\emph{The adjoins of ethiopia?}, whose associated logical form gives
the correct answer. Then there will be a dimension in the tensor
product for each pair of words; for example the dimensions associated
with (\emph{countries}, \emph{adjoins}) and (\emph{border},
\emph{adjoins}), as well as less useful pairs such as (\emph{5},
\emph{ethiopia}) would all have non-zero values in the tensor
product. Thus we are able to learn that if we see \emph{borders} in
the query, then a logical form whose canonical utterance contains the
term \emph{adjoins} is a likely candidate to answer the query.

\section{Empirical Evaluation}

\subsection{Dataset}

We evaluated our system on the \webquestions dataset
\cite{Berant:13}. This consists of 5,810 question-answer pairs. The
questions were obtained by querying the Google Suggest API, and
answers were obtained using Amazon Mechanical Turk. We used the
standard train/test split supplied with the dataset, and used
cross-validation on the training set for development purposes.

\subsection{Implementation}

We built our implementation on top of the \parasempre system
\cite{Berant:14}, and so our evaluation exactly matches theirs. Our
implementation is freely available online.\footnote{Location witheld
  to preserve anonymity.}  We substituted the paraphrase system of
\parasempre with our tensor kernel-based system (i.e.~we excluded
features from both the association and vector space models), but we
included the \parasempre features derived from logical forms.

To implement our tensor kernel of unigram features, we simply added
all pairs of terms in the query and canonical utterance as features;
in preliminary experiments we found that this was fast enough and we
did not need to use the kernel trick, which could potentially provide
further speed-ups. We did not implement any feature selection
methods which may also help with efficiency.

%% As features for
%% the query we use unigrams, and for the logical form we use the
%% automatically generated natural language gloss of the query generate
%% by \parasempre and extract unigrams in the same way as we do for the
%% query. We eliminate stopwords then apply the tensor kernel to generate
%% the vector for the instance.

For evaluation, we report the average of the F1 score measured on the
set of entities returned by the logical form when evaluated on the
database, when compared to the correct set of entities. This allows,
for example, to get a non-zero score for returning a similar set of
entities to the correct one. For example, if we return the set \{Jaxon
Bieber\} as an answer to the query \emph{Who is Justin Bieber's
  brother?}  we allow a nonzero score (the correct answer according to
the dataset is \{Jazmyn Bieber, Jaxon Bieber\}).

\subsection{Results}

Results are reported in Table \ref{table:results}. Our system achieves
an average F1 score of 40.1\%, compared to \parasempre's 39.9\%. Our
system runs faster however, due to the simpler method of generating
features. Evaluating using \parasempre on the development set took
22h31m; using the tensor kernel took 14h44m on a comparable machine.

Since we have adopted the logical form templates of \parasempre, our
upper bound or \emph{oracle F1 score} is the same, 63\%
\cite{Berant:14}. This is the score that would be obtained if we knew
which was the best logical form out of all those generated. In
contrast, Microsoft's \deepqa has an oracle F1 score of 77.3\%
\cite{Wang:14}; this could account for a large amount of the overall
increase in their system. There is no reported oracle score for the
Facebook system \cite{Bordes:14}.

\begin{table}[tb]
\centering
\begin{tabular}{l|c}
 & \textbf{Average F1 score} \\
\hline
\sempre \cite{Berant:13} & 35.7 \\
\parasempre \cite{Berant:14} & 39.9 \\
Facebook \cite{Bordes:14} & 41.8 \\
\deepqa \cite{Wang:14} & 45.3 \\
\hline
Tensor kernel with unigrams & 40.1 \\
\end{tabular}
\vspace{0.2cm}
\caption{Results on the \webquestions dataset, together with results
  reported in the literature.}
\label{table:results}
\end{table}

\section{Discussion}

Table \ref{table:features} shows the top unigram feature pairs after
training on the \webquestions training set. It is clear that, whilst
there are some superfluous features that simply learn to replace a
word with itself (for example \emph{currency} with \emph{currency},
there are obviously many useful features that would be nontrivial to
identify accurately. There are also spurious ones such as the pair
(\emph{live}, \emph{birthplace}); this is perhaps due to a large
proportion of people who live in their birthplace.

\begin{table}
\begin{minipage}[t]{.5\textwidth}
\centering
\csvreader[tabular=lc,
 table head=\textbf{Feature} &\textbf{Weight}\\\hline,
 late after line=\\]%
{unigramparams-short.csv}{}%
{(\csvcoli, \csvcolii) & \csvcoliii}%
\end{minipage}\begin{minipage}[t]{.5\textwidth}
\centering
\csvreader[tabular=lc,
 table head=\textbf{Feature} &\textbf{Weight}\\\hline,
 late after line=\\]%
{unigramparams-short-2.csv}{}%
{(\csvcoli, \csvcolii) & \csvcoliii}%
\end{minipage}
\vspace{0.2cm}
\caption{Top unigram pair features and their weights after training.}
\label{table:features}
\end{table}

In development, we found that ordering the training alphabetically by
the text of the query lead to a large reduction in
accuracy.\footnote{We omit the values since they were performed on an
  earlier version of our code and are not comparable.} Ordering
alphabetically when performing the split for cross validation (instead
of random ordering) means that a lot of queries on the same topic are
grouped together, increasing the likelihood that a query on a topic
seen at test time would not have been seen at training time. This
validates our hypothesis that simple techniques work well because of
the homogeneous nature of the dataset. We would argue that this does
not invalidate the techniques however, as it is likely that real-world
datasets also have this property.

It is a feature of our tensor product model that there is no direct
interaction between the features from the query and those from the
logical form. This is evidenced by the fact that the system has to
\emph{learn} that the term \emph{currency} in the query maps to
\emph{currency} in the canonical utterance. This hints at ways of
improving over our current system. More interestingly, it also means
that we are currently making very light use of the canonical utterance
generation; in the canonical utterance, \emph{currency} could be
replaced by any symbol and our system would learn the same
relationship. This points at another route of investigation involving
generating features for use in the tensor kernel directly from the
logical form instead of via canonical utterances.
 
\section{Conclusion}

We have shown semantic parsing via paraphrasing using unigram features
together with a tensor kernel performs comparably to more complex
systems on the \webquestions dataset. Our system is simpler to
implement and runs faster.

In future work, as well as looking at more sophisticated feature
inputs to the tensor kernel, we hope to work on improving the oracle
F1 score.

\bibliographystyle{splncs}
\bibliography{paper}
\end{document}